\documentclass[runningheads]{llncs}
\usepackage[T1]{fontenc}
\usepackage{graphicx}
\usepackage{xcolor}
\usepackage{multirow}

\begin{document}

\title{Adversarial Evasion Attack Efficiency against Large Language Models}

\titlerunning{DCAI 2024 preprint}

\author{Jo{\~{a}}o Vitorino\orcidID{0000-0002-4968-3653} \and
Eva Maia\orcidID{0000-0002-8075-531X} \and
Isabel Pra{\c{c}}a\orcidID{0000-0002-2519-9859}}

\authorrunning{ }

\institute{Research Group on Intelligent Engineering and Computing for Advanced Innovation and Development (GECAD), School of Engineering, Polytechnic of Porto (ISEP/IPP), 4249-015 Porto, Portugal\\
\email{\{jpmvo,egm,icp\}@isep.ipp.pt}}

\maketitle

\begin{abstract}
Large Language Models (LLMs) are valuable for text classification, but their vulnerabilities must not be disregarded. They lack robustness against adversarial examples, so it is pertinent to understand the impacts of different types of perturbations, and assess if those attacks could be replicated by common users with a small amount of perturbations and a small number of queries to a deployed LLM. This work presents an analysis of the effectiveness, efficiency, and practicality of three different types of adversarial attacks against five different LLMs in a sentiment classification task. The obtained results demonstrated the very distinct impacts of the word-level and character-level attacks. The word attacks were more effective, but the character and more constrained attacks were more practical and required a reduced number of perturbations and queries. These differences need to be considered during the development of adversarial defense strategies to train more robust LLMs for intelligent text classification applications.

\end{abstract}

\section{Introduction}
\label{sec:intro}

In recent years, the use of Large Language Models (LLMs) has grown exponentially, as they can help organizations automate a wide range of tasks. However, despite the benefits of using LLMs in artificial intelligence systems, they have vulnerabilities that must not be disregarded \cite{b6b5f50a}. When deployed for a text classifications task, an LLM may exhibit misclassifications if a text sample has subtle perturbations in some words or characters. This lack of robustness is one of the main challenges that undermine the trustworthiness of LLMs, and deep learning models in general, because they are susceptible to adversarial attacks \cite{Xu2020}.

LLM engineers and cybersecurity personnel still lack the tools to detect and prevent such perturbations in the large amounts of data that are used to train LLMs and in the queries that are then used by groups of users, so adversarial attacks pose a major threat to LLMs \cite{EuropeanUnionAgencyforCybersecurity2022a,SivaKumar2020}. To improve the security of LLMs, it is essential to methodically analyze and understand the impacts of the perturbations generated by different types of adversarial attacks, and assess if those attacks could be replicated by common users with a small amount of perturbations and a small number of queries to an LLM.

This work presents an analysis of the effectiveness, efficiency, and practicality of three different types of adversarial attacks against five different LLMs which are commonly used for text classification tasks. The models were used in a binary sentiment classification task with text samples representing positive and negative connotations. Several metrics were computed to analyze and compare each attack, considering number of misclassifications, the perturbations within an adversarial example and the computational cost of each attack.

The present paper is organized into multiple sections. Section \ref{sec:sota} provides a survey of previous work on adversarial attacks for text classification, with a focus on LLMs. Section \ref{sec:meths} describes the considered dataset, models, and attacks, as well as the computed evaluation metrics. Section \ref{sec:resu} presents and discusses the results of each attack in each metric. Finally, Section \ref{sec:conc} addresses the main conclusions and future research topics to be explored.

\section{Related Work}
\label{sec:sota}

Throughout the years, LLMs have been successfully applied to many different tasks across multiple domains \cite{10.1145/3641289}. However, these models are not flawless, so it is essential to carefully analyze their attack surface before deploying them to process the sensitive or confidential data of an organization \cite{b6b5f50a}.

Due to advances in chatbot technologies using LLMs, the scientific community is starting to pay more attention to the vulnerabilities that enable an attacker to perform prompt injection and exploit hidden backdoors, which are more common in such applications \cite{YAO2024100211}. However, since LLMs continue to be used for text classification applications, it is necessary to also take into consideration their vulnerabilities to adversarial attacks \cite{10.1145/3564281}.

An adversarial evasion attack can exploit an LLM to cause unexpected behaviors and significant disruptions in the business processes of an organization \cite{Vitorino2023}. By creating subtle perturbations in a text sample, an attacker can cause a model to misclassify that sample and assign it a completely different class with a high confidence score \cite{10.1145/3328795}. The perturbed text sample is designated as an adversarial example and can contain perturbations in a few words, or even in just a single word, making it almost imperceptible to a human \cite{Szegedy2014,ma2021exploiting}.

Even though all classification models are inherently susceptible to adversarial examples, different models will learn distinct simplifications and decision boundaries \cite{Goodfellow2015,10.1145/3564281}. Hence, some LLMs may be more vulnerable to certain perturbations than others, presenting model-specific edge cases that are hard to detect and address because the unexpected behaviors are only triggered by very specific expressions or word combinations \cite{zhang2019adversarial}.

To automate the process of finding these edge cases and exploiting an LLM, several adversarial evasion attack methods have been developed \cite{Xu2020}. Due to the different characteristics of the existing methods and diverse end goals of attackers, efforts are being made to systematize the possible attack vectors in the Adversarial Threat Landscape for Artificial-Intelligence Systems \cite{mitreatlas} knowledge base, and to complement it with case studies and demonstrations.

For a standard sentiment classification task, most of the attacks targeting LLMs generate word-level perturbations to replace one or more words with other similar words or expressions, attempting to obfuscate the connotation of a text sample \cite{li2020bertattack,ribeiro2020accuracy,jia-etal-2019,Li2019}. On the other hand, some approaches generate character-level perturbations to slightly modify the characters of a word by adding, removing, or switching them with other similar characters, attempting to exploit the tokenization, stemming, or lemmatization processes of a model \cite{pruthi2019combating,9833641}.

Since these different types of attacks can lead to very different misclassifications, and may not be efficient or practical in every deployment scenario, it is important to perform a methodical evaluation of their impact on the commonly used LLMs. To the best of our knowledge, no previous work has compared the efficiency and practicality of the three considered attack methods for a sentiment classification task with the five considered LLMs.

\section{Methods}
\label{sec:meths}

This work was carried out on a machine with an NVIDIA Quadro P4000, which follows the Pascal GPU microarchitecture, has 8 GB GDDR5 memory and a total of 1792 CUDA cores. The implementation relied on the Python 3 programming language and the following libraries: textattack for the implementation of the adversarial attacks, huggingface for the loading of the dataset and the models, and pytorch to compute their predictions.

\subsection{Dataset and Models}

The utilized dataset for the sentiment classification task was RottenTomatoes \cite{pang2005seeing}, a publicly available dataset of movie reviews that is commonly used to benchmark the performance of language models for sentiment analysis. It is a balanced dataset with 10662 text samples of full sentences of numerous reviews, with half of them being assign to a positive connotation label and the other half to a negative connotation label.

The selected models for the analysis were BERT, RoBERTa, DistilBERT, ALBERT, and XLNet, because they presented specific differences that may affect the effectiveness and practicality of an attack. These are all transformer-based models commonly used for diverse natural language processing tasks, including sentiment classification. The models used in this work have been pre-trained for the specific task of sentiment classification of movie reviews, having their own fine-tuned tokenizer already available at the huggingface library.

BERT \cite{devlin2019bert} is an autoencoding model that has been well-established in the scientific community because it has a very good performance in discerning the context of words by using random masking of the tokens of a sentence. RoBERTa \cite{liu2019roberta} is a variant of BERT that has more computational capacity and an improved training methodology with a higher quantity of training data and a higher training length, attempting to provide better performance than BERT.

On the other hand, both DistilBERT \cite{sanh2020distilbert} and ALBERT \cite{lan2020albert} are smaller and more lightweight versions of BERT, designed to provide faster predictions and be more efficient while attempting to preserve a similar performance, making the benefits of BERT more accessible for smaller organizations with a lower computational capacity. In turn, XLNet \cite{yang2020xlnet} is an autoregressive model, but it is based on similar concepts as the previous autoencoding models and uses permutation-based training to permute the tokens of a sentence. 

\subsection{Attacks and Metrics}

Three types of adversarial evasion attack methods were selected because they follow different approaches to generate adversarial examples, leading to distinct impacts on the robustness of the models. The BERTAttack \cite{li2020bertattack} was used because it employs its own BERT model to generate word-level perturbations and attack another LLM, effectively placing two LLMs competing against each other. The ChecklistAttack \cite{ribeiro2020accuracy} was also utilized because it performs similar word-level perturbations, but relies on a constrained data generation process to only replace words with other suitable options that are available in a predefined checklist. The TypoAttack \cite{pruthi2019combating} attack is substantially different, as it performs more specific character-level perturbations by inserting characters, deleting characters, swapping neighboring characters within a word, or swapping characters for adjacent keys on a standard QWERTY keyboard.

The standard evaluation metric to evaluate and compare the effectiveness of the adversarial attacks is the Misclassification Rate (MR). It measures the proportion of text samples that were misclassified, reflecting the effectiveness of an attack in deceiving a model. Since its value is increased from 0 to 100\% as more adversarial examples are misclassified, a low MR indicates that an LLM is more robust against a certain attack.

To evaluate the efficiency and practicality of an attack, it was necessary to consider the main characteristics of how that attack is done, which are not usually considered in simple performance evaluations. To compare the amount of perturbations each attack method generated and its computational cost, two evaluation metrics were computed: Average Perturbed Words (APW) and the Average Required Queries (ARQ).

The APW measures the number of modified words within a perturbed text sample, computing the average of all perturbed text samples. Even though the character-level perturbations did not replace entire words, these were still counted as perturbed words because they were modified. In turn, the ARQ measures the number of attempts that queried a model's class predictions until the right perturbations were found and the text sample was misclassified, computing the average of all perturbed text samples.

\section{Results and Discussion}
\label{sec:resu}

An effectiveness, efficiency, and practicality evaluation was performed by using each method to perform adversarial evasion attacks against each LLM. The considered metrics, MR, APW, and ARQ, were computed throughout the attacks, and the results of each method were analyzed and compared. Table \ref{tab_res} presents a summary of the obtained results.

\setlength{\tabcolsep}{10pt}
\renewcommand{\arraystretch}{1.2}

\begin{table*}[h]
\centering
\caption{\label{tab_res}
Obtained results for the sentiment classification task.
}
\begin{tabular}{lllll}
\hline
\textbf{Model} &
\textbf{Attack} &
\textbf{MR (\%)} &
\textbf{APW (\#)} &
\textbf{ARQ (\#)} \\
\hline
\multirow{3}{5em}{BERT} & BERTAttack & 100 & 3.09 & 135.17 \\
& ChecklistAttack & 40  & 1.06 & 2.17 \\
& TypoAttack & 50 & 1.58 & 381.10 \\
\hline
\multirow{3}{5em}{RoBERTa} & BERTAttack & 80 & 3.14 & 104.5 \\
& ChecklistAttack & 20 & 1.15 & 2.25 \\
& TypoAttack & 50 & 1.82 & 356.62 \\
\hline
\multirow{3}{5em}{DistilBERT} & BERTAttack & 100 & 3.08 & 104.86 \\
& ChecklistAttack & 30 & 1.12 & 2.10 \\
& TypoAttack & 50 & 1.64 & 355.20 \\
\hline
\multirow{3}{5em}{ALBERT} & BERTAttack & 90 & 2.48 & 98.67 \\
& ChecklistAttack & 10 & 1.09 & 2.11 \\
& TypoAttack & 80 & 1.70 & 339.15 \\
\hline
\multirow{3}{5em}{XLNet} & BERTAttack & 100 & 2.43 & 85.33 \\
& ChecklistAttack & 10 & 1.08 & 2.20 \\
& TypoAttack & 90 & 1.62 & 362.89 \\
\hline
\end{tabular}
\end{table*}

The adversarial examples generated by the three different types of attacks had very distinct impacts. The word-level BERTAttack and the character-level TypoAttack obtained very high misclassification rates, which demonstrated that both types of perturbations can be used to attack LLMs. Despite ChecklistAttack also being a word-level attack, it was only able to cause a maximum of 40\% of misclassifications, possibly due to its more constrained data generation process that did not freely modify any word in a text sample. Since it only modified one or two words with similar ones from a predefined checklist, the attack didn't transfer well to the dataset used in this work.

The most effective attack against the considered models was BERTAttack. It achieved a misclassification rate of approximatelly 100\% against BERT, DistilBERT, and XLNet, so the perturbations were able to deceive the models into misclassifying almost every text sample. Furthermore, it reached 80\% against RoBERTa and 90\% against ALBERT, which evidenced the high effectiveness of using an LLM to attack another LLM. Nonetheless, it completely replaced two, three, or four words of a text sample, as can be seen in Fig. \ref{fig_res_1}. These replacements sometimes made the text sample lose some coherence, which would make the attack less realistic and more evident for human personnel overseeing the model's behavior.

\begin{figure*}[ht]
\centering
\fbox{%
\begin{minipage}{35em}
\centering

Misclassification: \textbf{\textcolor{green!70!olive}{Positive}} (100\% confidence) ---> \textbf{\textcolor{orange!70!red}{Negative}} (80\% confidence)

\hfill

"the story gives ample \textbf{\textcolor{green!70!olive}{opportunity}} for large-scale action and \textbf{\textcolor{green!70!olive}{suspense}}, which director shekhar kapur supplies with \textbf{\textcolor{green!70!olive}{tremendous}} skill."

"the story gives ample \textbf{\textcolor{orange!70!red}{need}} for large-scale action and \textbf{\textcolor{orange!70!red}{screenplay}}, which director shekhar kapur supplies with \textbf{\textcolor{orange!70!red}{most}} skill."

\end{minipage}
}%
\caption{\label{fig_res_1}
Misclassification caused by three word-level perturbations.
}
\end{figure*}

On the other hand, the TypoAttack only modified a few characters in one or two words of a text sample. It was able to cause 50\% of misclassifications in the bigger and more complex LLMs like BERT and RoBERTa, and even reached 80\% against the smaller ALBERT and up to 90\% against XLNet. The slight modifications can be harder to notice because human personnel usually overlooks missing letters or similar letters like "i" and "l", as can be seen in Fig. \ref{fig_res_2}. These perturbations can go unnoticed in a dataset and even lead to the creation of hidden backdoors in a model, so they pose a very concerning security risk for organizations that use large amounts of unverified data from unreliable sources to train LLMs.

\begin{figure*}[ht]
\centering
\fbox{%
\begin{minipage}{35em}
\centering

Misclassification: \textbf{\textcolor{green!70!olive}{Positive}} (83\% confidence) ---> \textbf{\textcolor{orange!70!red}{Negative}} (74\% confidence)

\hfill

"lovingly photographed in the manner of a golden book sprung to life, stuart little 2 manages sweetness largely \textbf{\textcolor{green!70!olive}{without}} stickiness."

"lovingly photographed in the manner of a golden book sprung to life, stuart little 2 manages sweetness largely \textbf{\textcolor{orange!70!red}{wlthout}} stickiness."

\end{minipage}
}%
\caption{\label{fig_res_2}
Misclassification caused by one character-level perturbation.
}
\end{figure*}

Nonetheless, despite all the considered models being susceptible to adversarial examples with both word-level and character-level perturbations, a large number of queries was required to attack them. To generate an example that actually deceived a model, BERTAttack required from 85 to over 135 queries, whereas TypoAttack required the substantially higher number of 355 to over 381 queries, on average. This limits the functionality and practicality of such attacks, as an attacker may not be able to perform such a high number of similar queries to one single model to try to generate a working example. Therefore, more constrained attacks like ChecklistAttack may not be able to cause many misclassifications, but they may be more practical and efficient against an LLM deployed in a real system.

Overall, the results of these three different types of attacks suggest that their drawbacks may outweigh their benefits in a real deployment scenario. To perform a realistic adversarial evasion attack that is both effective and practical against a real system, it will be necessary to achieve a good balance between the considered metrics. The performed attacks should be able to cause as many misclassifications as possible, while generating a minimal number of perturbations and performing just a few queries to a model.

\section{Conclusion}
\label{sec:conc}

This work analyzed and compared the effectiveness, efficiency, and practicality of three different types of adversarial evasion attacks to provide a better understanding of the impacts of different types of perturbations. The attacks were performed in a sentiment classification task with the text samples of movie reviews of the RottenTomatoes dataset. The analysis included the amount and the quality of perturbations of each attack, their computational efficiency, and also the susceptibility of five different LLMs.

The considered BERTAttack, ChecklistAttack, and TypoAttack, were performed against BERT, RoBERTa, DistilBERT, ALBERT, and XLNet models. The first, BERTAttack, achieved the best overall misclassification rates, although it required many queries until it generated word-level perturbations that could actually deceive a model, and those were not always fully coherent. TypoAttack was able to generate the smallest and less noticeable character-level perturbations while still causing many misclassifications, but the substantially higher number of required queries could limit its practicality against a deployed LLM. ChecklistAttack required a very small number of queries in comparison with the other two types of attacks, although the constrained number of perturbed words reduced its effectiveness.

The attacks had very distinct impacts, which need to be considered when using them in a real deployment scenario, and should also be considered by ML engineers and cybersecurity personnel when creating LLMs. The obtained results highlight a potentially valuable adversarial defense strategy: automatically verifying a sequence of queries during the inference phase to check if there are too many similar queries being used. If this verification was performed in real-time, it would be possible to temporarily block such suspiciously similar queries, signaling them for a more in-depth analysis by cybersecurity personnel. This would work as a safeguard to prevent the LLM from being exploited by groups of users with malicious intent.

In the future, it is pertinent to further contribute to robustness research by replicating this analysis with other types of adversarial attacks and perform them against more LLMs in text classifications tasks. As the threat of attacks against LLMs increases, it is also important to explore the possible defenses against further attacks like prompt injections, address privacy and confidentiality concerns, and investigate the transferability of adversarial attacks and defense strategies techniques across different natural language processing tasks.

\begin{credits}
\subsubsection{\ackname} This work was supported by the CYDERCO project, which has received funding from the European Cybersecurity Competence Centre under grant agreement 101128052. This work has also received funding from UIDB/00760/2020.
\end{credits}

\bibliographystyle{splncs04}
\bibliography{mybibliography}

\end{document}